\documentclass[compactaffiliation]{Interspeech}
\interspeechcameraready

\usepackage{cite}
\usepackage{amsmath,amssymb,amsfonts}
\usepackage{algorithmic}
\usepackage{graphicx}
\usepackage{textcomp}
\usepackage{xcolor}
\usepackage{url}
\usepackage{multirow}
\usepackage[normalem]{ulem}
\usepackage{subcaption}

\def\BibTeX{{\rm B\kern-.05em{\sc i\kern-.025em b}\kern-.08em
    T\kern-.1667em\lower.7ex\hbox{E}\kern-.125emX}}
\begin{document}

\newcommand{\fv}[1]{\textcolor{blue}{[FV: #1]}}
\newcommand{\fvdel}[1]{\textcolor{blue}{\sout{[FV: #1]}}}

\newcommand{\fc}[1]{\textcolor{red}{[FC: #1]}}
\newcommand{\fcdel}[1]{\textcolor{red}{\sout{[FC: #1]}}}

\newcommand{\spadd}[1]{\textcolor{green}{[SP: #1]}}
\newcommand{\spdel}[1]{\textcolor{green}{\sout{[SP: #1]}}}

\newcommand{\pmadd}[1]{\textcolor{purple}{[PM: #1]}}
\newcommand{\pmdel}[1]{\textcolor{purple}{\sout{[PM: #1]}}}

\title{{How to Connect Speech Foundation Models and Large Language Models?\\ What Matters and What Does Not}}

\author[affiliation={1,3}, equalcontribution]{Francesco}{Verdini}
\author[affiliation={1,2,3}, equalcontribution]{Pierfrancesco}{Melucci}
\author[affiliation={2,3}, equalcontribution]{Stefano}{Perna}
\author[affiliation={3,4}, equalcontribution]{Francesco}{Cariaggi}
\author[affiliation={5}]{Marco}{Gaido}
\author[affiliation={5}]{Sara}{Papi}
\author[affiliation={6,7}]{Szymon}{Mazurek}
\author[affiliation={7}]{Marek}{Kasztelnik}
\author[affiliation={5}]{Luisa}{Bentivogli}
\author[affiliation={3,4}]{Sébastien}{Bratières}
\author[affiliation={2}]{Paolo}{Merialdo}
\author[affiliation={1}]{Simone}{Scardapane}

% Affiliations

\affiliation{}{Sapienza University of Rome}{Italy}
\affiliation{}{Roma Tre University}{Italy}
\affiliation{}{Translated}{Italy}
\affiliation{}{Pi School}{Italy}
\affiliation{}{Fondazione Bruno Kessler}{Italy}
\affiliation{}{AGH University of Krakow}{Poland}
\affiliation{}{ACC Cyfronet AGH}{Poland}

% Emails
\email{francesco.verdini@uniroma1.it, pierfrancesco.melucci@uniroma1.it, stefano.perna@uniroma3.it, francesco.cariaggi@picampus-school.com}


\setlength\titlebox{5.75cm}
\maketitle
\begin{abstract}
The remarkable performance achieved by Large Language Models (LLM) has driven research efforts to leverage them for a wide range of tasks and input modalities. In speech-to-text (S2T) tasks, the emerging solution consists of projecting the output of the encoder of a Speech Foundational Model (SFM) into the LLM embedding space through an adapter module. However, no work has yet investigated how much the downstream-task performance depends on each component (SFM, adapter, LLM) nor whether the best design of the adapter depends on the chosen SFM and LLM. To fill this gap, we evaluate the combination of 5 adapter modules, 2 LLMs (Mistral and Llama), and 2 SFMs (Whisper and SeamlessM4T) on two widespread S2T tasks, namely Automatic Speech Recognition and Speech Translation. Our results demonstrate that the SFM plays a pivotal role in downstream performance, while the adapter choice has moderate impact and depends on the SFM and LLM.
\end{abstract}

\section{Introduction}
\label{sec:intro}

\begin{figure}[t]
    \centering
    \includegraphics[width=1\linewidth]{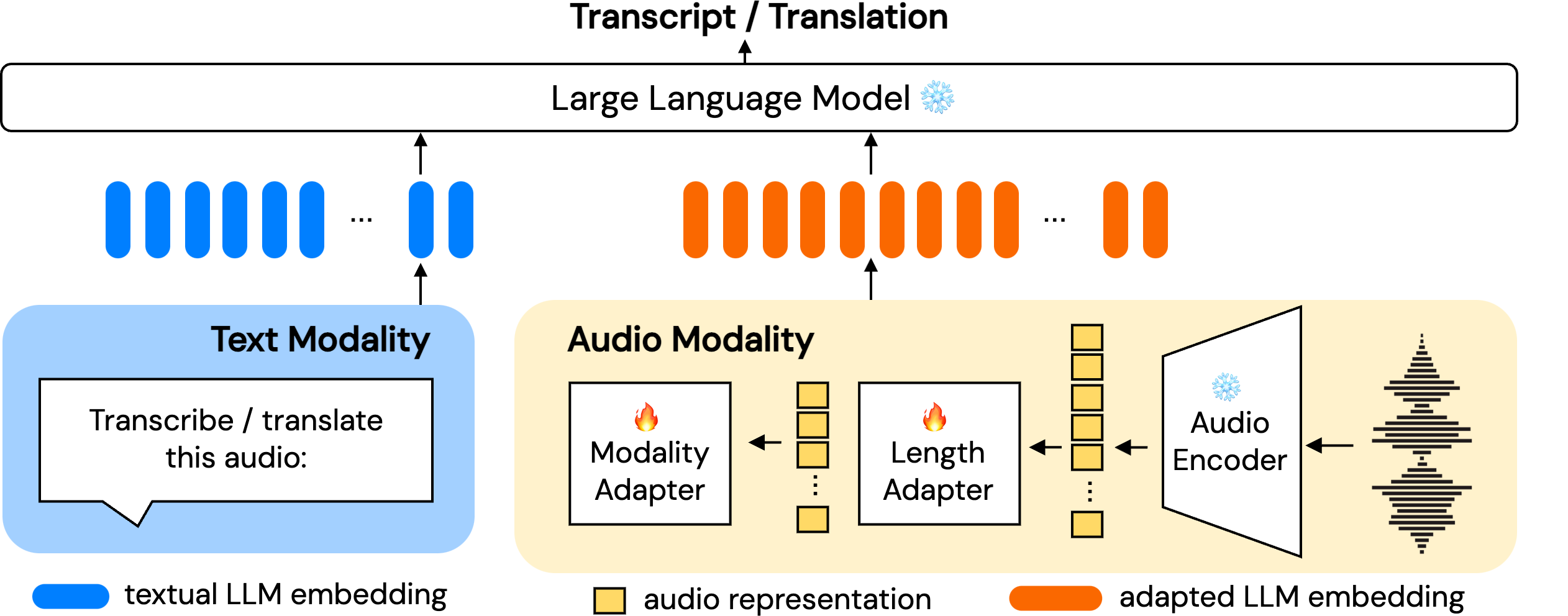}
    \caption{Schema of the SFM+LLM architecture.}
    \label{fig:sfm+llm}
\end{figure}

The success of Large Language Models (LLMs) \cite{llama_3_1} has attracted significant interest in extending their capabilities to handle various input modalities such as vision \cite{llava} and speech~\cite{salmonn}. In the speech scenario, several studies~\cite{salmonn,speech_llama,wang2023blsp,hu2024wavllm,wang2024blsp} have proposed the integration of a pretrained Speech Foundation Model (SFM) encoder with a pretrained LLM through an \textit{adapter} module, realizing the SFM+LLM new architectural paradigm \cite{sfm_survey}. The adapter can be decomposed into two components, as shown in Figure \ref{fig:sfm+llm}: a \textit{length adapter}, which compresses the speech sequence along the time dimension, and a \textit{modality adapter}, which maps the compressed input into an embedding space compatible with the LLM. The SFM+LLM solution exploits, on the one hand, the SFM ability to extract high-quality semantic representations of the speech input and, on the other, the fluency and vast linguistic knowledge of LLMs, achieving competitive scores for widespread tasks such as Automatic Speech Recognition (ASR) \cite{yu2023connectingspeechencoderlarge} and Speech Translation (ST)~\cite{wang2024blsp}.

However, being a recent trend, 
%the parallel 
research efforts have mostly been devoted to demonstrating the effectiveness of this paradigm over traditional methods, without delving into the single design choices. Specifically, many architectural solutions have been proposed for the length adapter, often employed to both reduce the LLM computational costs 
%(quadratic with the input length) 
and the modality mismatch with the textual sequences. These methods span from fixed downsampling, obtained either with a stack of strided convolutions \cite{yu2023connectingspeechencoderlarge} or with window-level Q-Former \cite{salmonn}, to modules with variable compression rates that reduce the input sequence based on its semantic content, such as Continuous Integrate-and-Fire (CIF) \cite{cif} and CTC compression~\cite{ctc_compression}.
Nonetheless, a comprehensive study on the adapter choice is missing: while some comparisons are present in the literature~\cite{wang2023blsp,wang2024blsp,embarrassinglysimple}, these evaluations are narrow in scope and assume the optimal solution is independent of the chosen SFM or LLM.

In this work, we explore whether a one-size-fits-all design choice for the adapter exists that maximizes ASR and ST performance or if this depends on the selected SFM and LLM. Through a systematic comparison of a wide range of adapters proposed in the literature and by analyzing their impact in combination with widely used SFMs (Whisper \cite{whisper} and SeamlessM4T \cite{seamless}) and LLMs (Llama \cite{llama_3_1} and Mistral \cite{mistral}), our contributions can be summarized as follows:
\begin{itemize}
    \item We prove that performance highly varies when a different SFM is selected (on average, \textgreater 2 COMET points for ST and 1 WER for ASR), whereas the choice of the LLM and adapter has a less pronounced impact on the final performance.

    \item We show that there is no one-size-fits-all solution for the adapter as its choice highly depends on the selected SFM and LLM combination.
    %, suggesting that reducing the length mismatch between speech and textual representation is useful in terms of computational costs but does not significantly impact the overall output quality.
\end{itemize}
Our codebase will be released under the Apache 2.0 License upon paper acceptance.

\section{Methodology}
\label{sec:method}

% These components interact to generate the target translation $y$ from the input audio $x$ in the following way:
% 
% \begin{equation}
%     y = \mathcal{L} \left( \left[\mathcal{L}_E(\text{``can you transcribe''}), \, \mathcal{A}\left( \mathcal{S}_E(x) \right) \right] \right)
% \end{equation}
% 
% 
% where $[\cdot, \cdot]$ refers to the concatenation operator and

Figure \ref{fig:sfm+llm} illustrates the design of SFM+LLM solutions, highlighting the three key components: the SFM encoder, the adapter, and the LLM. The processing pipeline begins with an input audio segment $a$, which is first encoded by the SFM encoder $\mathcal{S}_E$ into a sequence of learned audio representations. These are then processed by the adapter $\mathcal{A}$ and the resulting embeddings are concatenated to the embeddings $\mathcal{P}$ of a task-specific textual prompt.\footnote{In our experiments, we use as textual prompt \textit{``can you transcribe $L_s$?''} for ASR, where $L_s$ is the language of the audio, and \textit{``can you translate from $L_s$ to $L_t$?''} for ST, where $L_t$ is the target language.} The joining operation $J$, which combines the audio and prompt embeddings, follows the same methodology established in the framework proposed by LLaVa \cite{llava} to construct the final prompt that is passed to the LLM. In particular, with $\mathcal{L}$ as the LLM, we can formally express the generation of the output transcription or translation $y$ through the following equation:
\begin{equation}
y = \mathcal{L}\left(J\left(\mathcal{P}, \mathcal{A}\left( \mathcal{S}_E(a) \right)\right)\right)
\end{equation}

As the main rationale behind the SFM+LLM solution comes from the possibility of training a high-quality ASR or ST system without large training datasets -- thus with limited computational costs and memory requirements -- we keep $\mathcal{L}$ and $\mathcal{S}_E$ frozen, training only $\mathcal{A}$. This solution is coherent with previous work \cite{wang2023slm}, which showed that the gains obtained by fine-tuning the whole SFM encoder and LLM do not justify the additional costs when training models at this scale and for these tasks.
Within this framework, we answer our research question on the relative importance of the three components for the downstream performance by varying each of them as illustrated in the following sections.

\subsection{SFM Encoder}

To investigate the impact of the SFM encoder, we use two widely recognized SFMs for speech representation extraction: Whisper~\cite{whisper} and SeamlessM4T~\cite{seamless}. In particular, we use the large version of both of them, namely Whisper large-v3,\footnote{\url{https://huggingface.co/openai/whisper-large-v3}} and SeamlessM4T v2-large.\footnote{\url{https://huggingface.co/facebook/seamless-m4t-v2-large}} While Whisper is the most popular SFM in recent works on SFM+LLM \cite{salmonn,speech_llama,qwen2_audio,llast,zerost}, representing a natural choice, the usage of SeamlessM4T has never been explored to the best of our knowledge. Both models have demonstrated strong performance in multilingual speech processing tasks, with Whisper showing particular strength in ASR tasks and SeamlessM4T excelling in speech-to-speech and speech-to-text translation. Moreover, their architectural differences make them interesting candidates for comparative analysis in our study. Nonetheless, we opted for it not only for its recognized quality, but also because its design is very different from Whisper. SeamlessM4T is built with a customized version of Conformer layers \cite{gulati20_interspeech} instead of Transformer ones, and also the compression factor of the input sequence is very different.
While both models process audio sequences where each frame represents 10ms of audio (100 frames per second), they differ in their temporal resolution: Whisper emits one frame every 20ms (50 frames per second, 2× downsampling), while SeamlessM4T's encoder returns one frame every 160ms (6.25 frames per second, 16× downsampling). This substantial difference in temporal resolution, with SeamlessM4T performing much more aggressive downsampling, provides an interesting opportunity to study how these architectural choices impact the optimal adapter design.

\begin{figure*}
    \centering
    \begin{subfigure}[t]{0.13\textwidth}
            \centering
            \caption{Base}
            \vspace{15pt}
            \hspace*{9pt}\includegraphics[width=0.85\linewidth,  keepaspectratio]{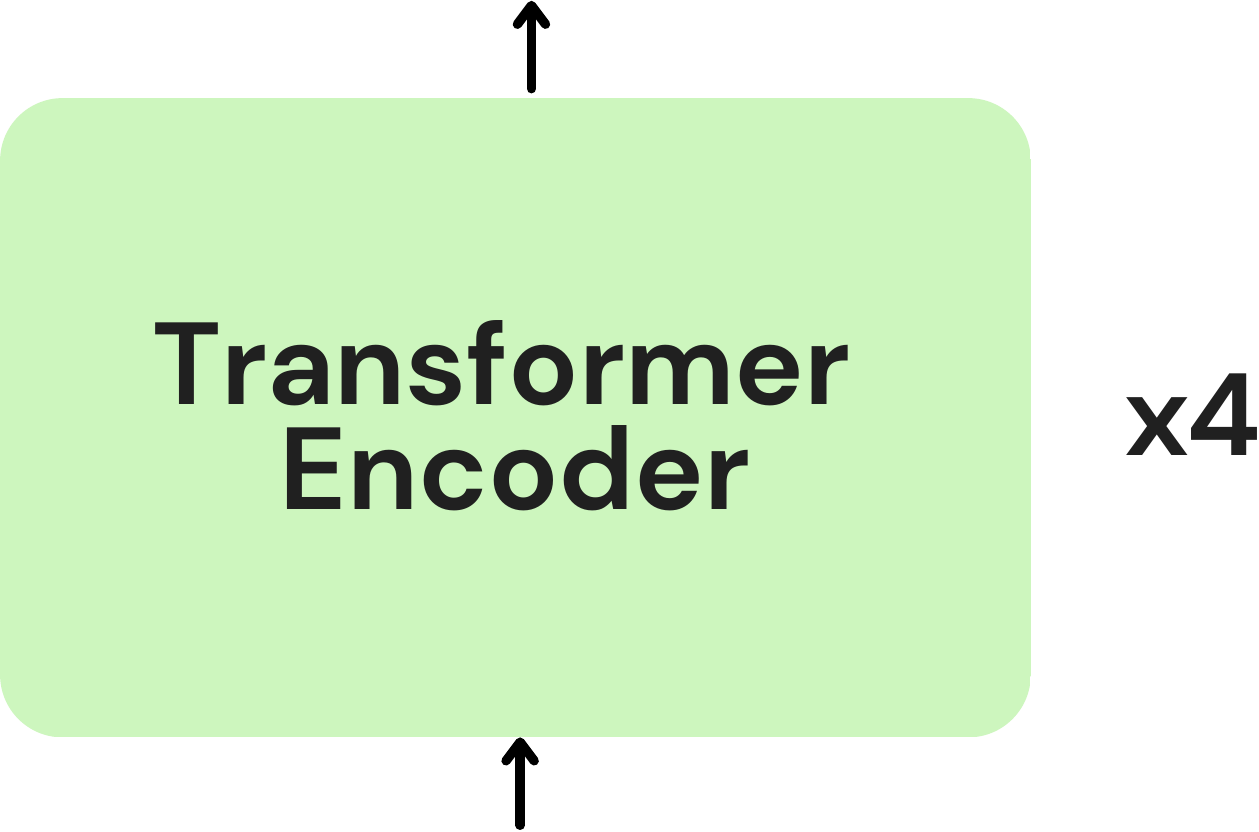}
        \end{subfigure}
        \hfill
        \begin{subfigure}[t]{0.18\textwidth}
            \centering
            \caption{Conv-based}
            \hspace*{10pt}\includegraphics[width=0.8\linewidth, keepaspectratio]{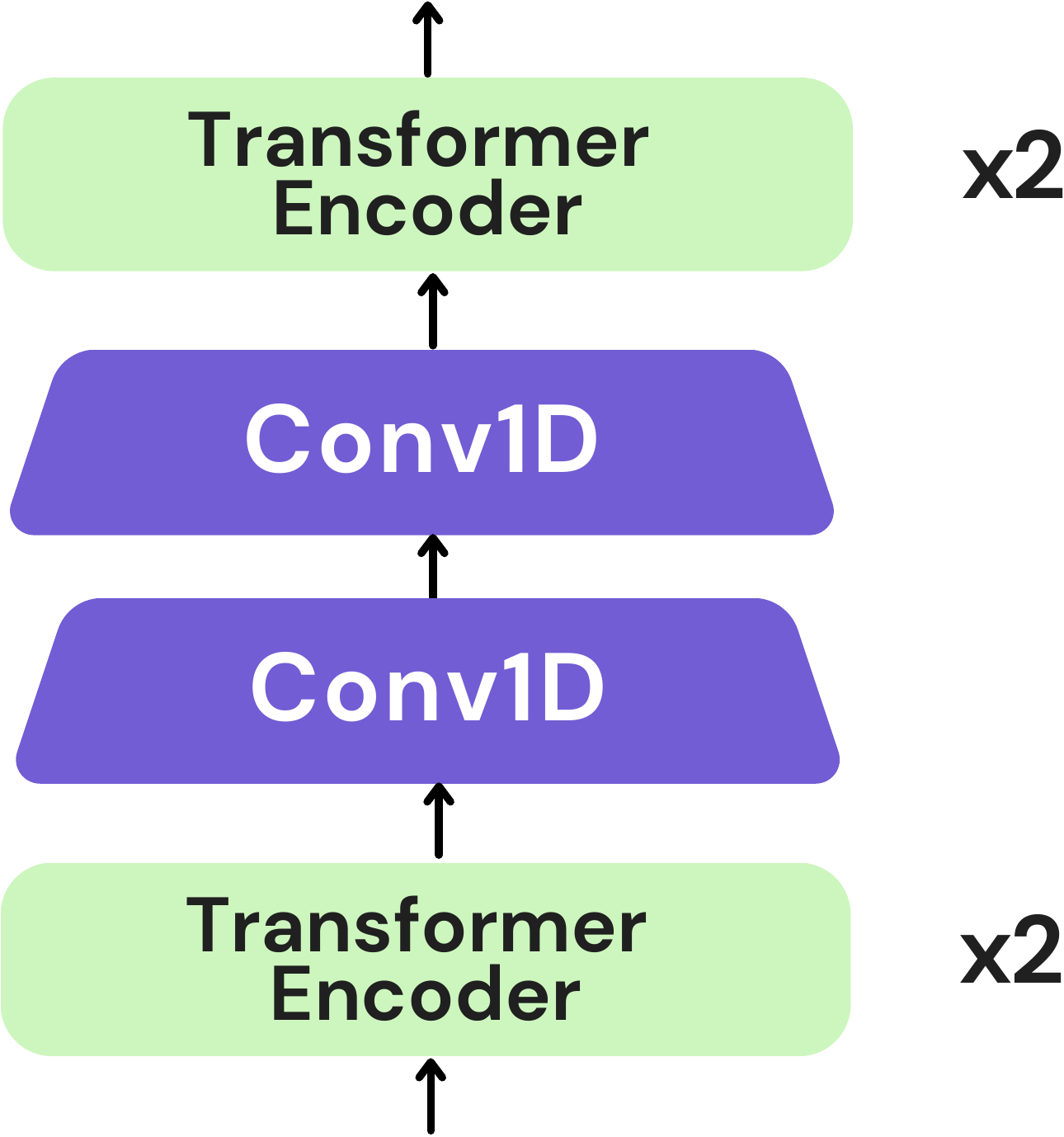}
        \end{subfigure}
        \hfill
        \begin{subfigure}[t]{0.2\textwidth}
            \centering
            \caption{CIF-based}
            \includegraphics[width=\linewidth,  keepaspectratio]{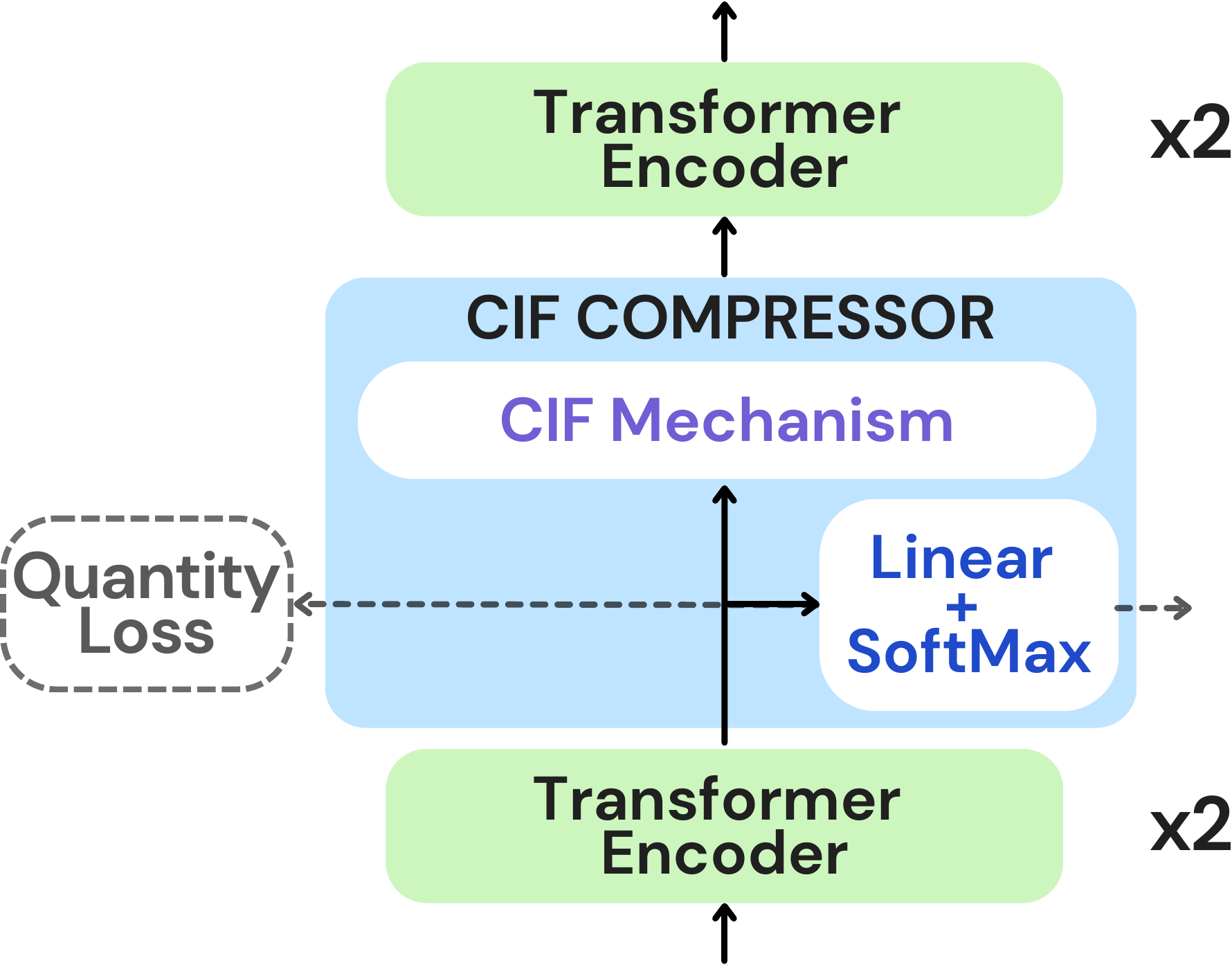}
        \end{subfigure}
        \hfill
        \begin{subfigure}[t]{0.19\textwidth}
            \centering
            \caption{CTC-based}
            \hspace*{10pt}\includegraphics[width=0.85\linewidth,  keepaspectratio]{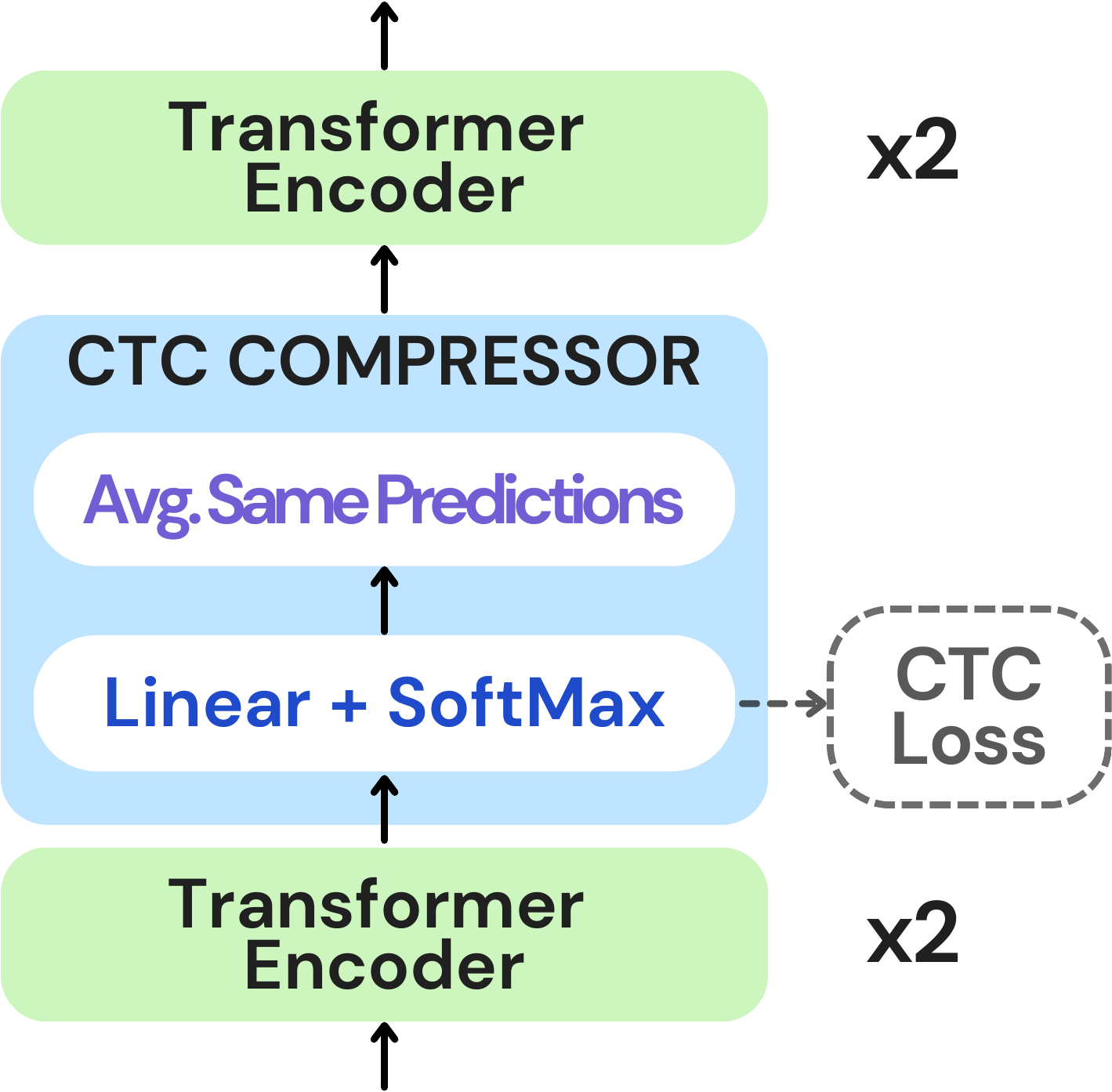}
        \end{subfigure}
        \hfill
        \begin{subfigure}[t]{0.24\textwidth}
            \centering
            \caption{WLQ-former}
            \vspace{2pt}
            \includegraphics[width=\linewidth,  keepaspectratio]{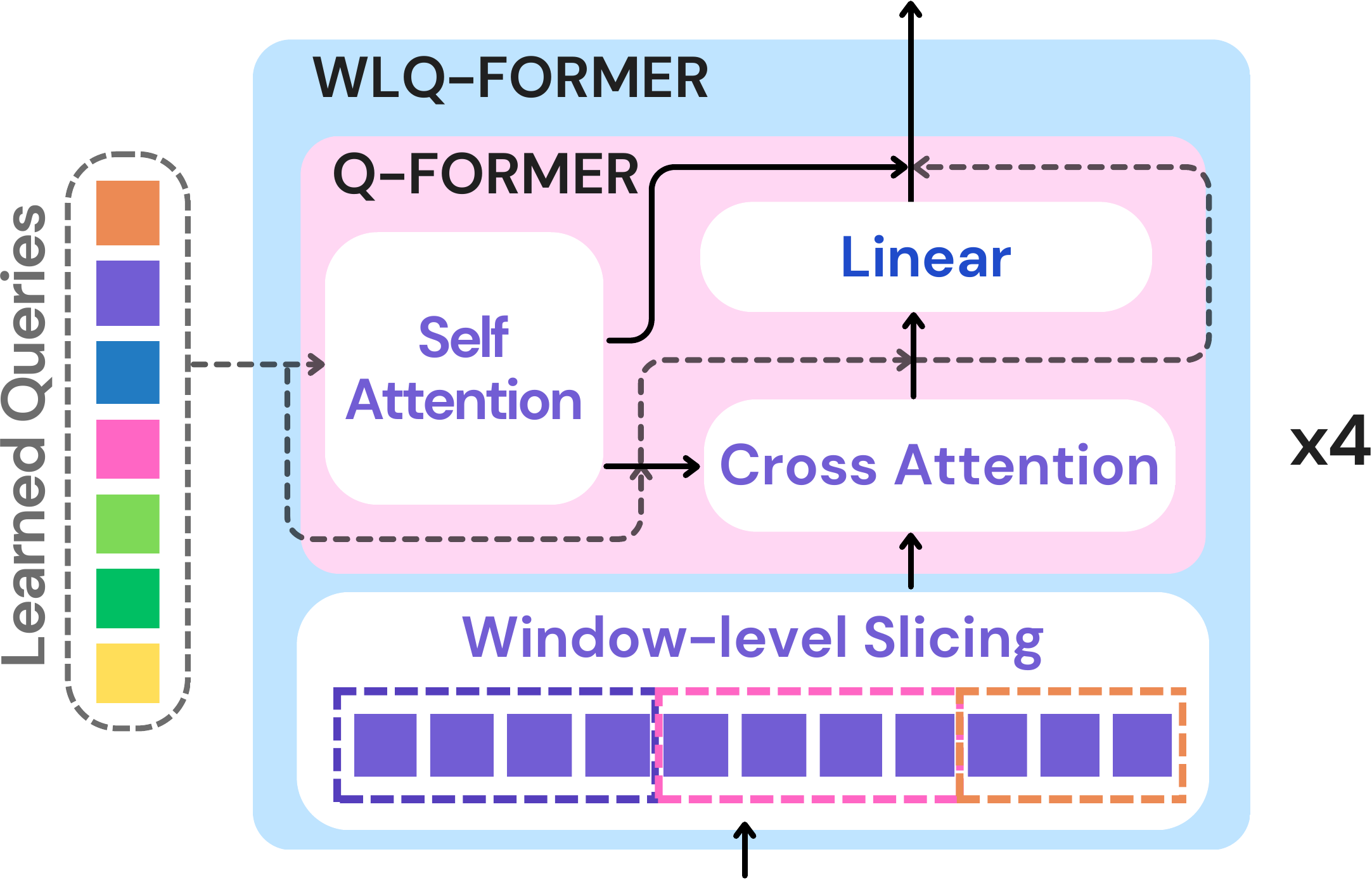}
        \end{subfigure}
    \caption{Representation of the adapters analyzed in the paper.}
    \label{fig:adapters}
\end{figure*}

\subsection{Adapter}

As we keep the SFM and LLM frozen, we design adapters with high representation capacity, allowing for an effective mapping of the embeddings to the LLM input space. To do so, we follow \cite{wang2023slm,speech_llama} which use a stack of vanilla Transformer \cite{transformer} encoder layers with bidirectional self-attention as modality adapters and investigate different methods as adapters. The adapters are trained using a cross-entropy loss on the output of the LLM having the transcripts for ASR and translations for ST as target, unless stated otherwise. One key characteristic of adapters is how they handle the sequence length. Some, like the Base adapter, keep the original one, while others reduce it using different strategies. Conv-based and WLQ-former-based adapters apply a constant-factor compression, while CIF-based and CTC-based adapters use a content-based one. Overall, we investigate the following 5 adapters, shown in Figure \ref{fig:adapters}.

\noindent
\textbf{Base.} We use 4 transformer encoder layers with hidden size 768, with 3072 as the intermediate size and using 12 attention heads. This adapter does not perform any length adaptation.

\noindent \textbf{Conv-based.} We introduce 2 1D-convolutional layers with stride 2 after the second layer of the Base adapter, resulting in a final compression factor of 4.
% The Base adapter is extended by introducing 2 convolutional layers with stride 2 after the second layer. 
No auxiliary loss is used. %The final compression factor is 4.

\noindent
\textbf{CIF-based.} Similarly to Conv-based, 
%The \textbf{Base} 
the adapter is extended by introducing a Continuous Integrate-and-Fire (CIF) \cite{cif} length adapter after the second Transformer layer. CIF is a sequence compression mechanism that accumulates input features over time and emits an output when a given integration threshold is reached, enabling variable-length sequence compression while preserving key information. To train this module, we add two auxiliary losses: a Connectionist Temporal Classification (CTC) loss \cite{graves2006ctc} with the transcripts as target, following \cite{multitask_asr}, and a quantity loss that controls the compression factor. The weight associated with both auxiliary losses is 0.1. On average, this corresponds to a compression factor of 3 with SeamlessM4T and 25 with Whisper.

\noindent
\textbf{CTC-based.} In this case, the length adapter is a CTC-based compression \cite{ctc_compression}, which collapses consecutive equal predictions of a CTC module by averaging the corresponding vectors, trained on the transcripts with an auxiliary CTC loss as done in CIF-based. On average, this corresponds to a compression factor of 2 for SeamlessM4T and 13 for Whisper.

\noindent
\textbf{WLQ-former.} This adapter performs both modality and length adaptation with a window-level Q-Former \cite{salmonn}. This module processes variable-length encoded speech sequences by dividing them into fixed-length windows of encoded frames and feeding each of these non-overlapping windows to a Q-former architecture \cite{blip2}. The Q-former uses a fixed and configurable number of learnable query vectors to attend to each window through cross-attention. As a result, the compression factor is controlled by the window length and the number of queries, which we set to 0.33 seconds and 1 respectively as per \cite{salmonn}, and therefore results in 2 for SeamlessM4T and 16 for Whisper.

\begin{table}[ht]
\setlength{\tabcolsep}{1.5pt}
\centering
\caption{Compression rate for each configuration of SFM/Adapter.}
\label{tab:compression_ratio}
\begin{tabular}{c|l|c|c}
\hline
\textbf{SFM}                        & \textbf{Adapter} & \textbf{Compression ratio}  & \textbf{Sampling rate (Hz)}\\
\hline
\multirow{5}{*}{\rotatebox{90}{Seamless}} & Base                         & 1:1 & 6.25                     \\
                                    & CIF-based                           & 3:1 & 2.08                    \\
                                    & Conv-based                          & 4:1 & 1.56                    \\
                                    & CTC-based                           & 2:1 & 3.12                   \\
                                    & WLQ-former                      & 2:1 & 3.12                     \\ 
\hline
\multirow{6}{*}{\rotatebox{90}{Whisper}}  & Base                         & 1:1  & 50.00                   \\
                                    & CIF-based                           & 25:1  & 2.00                   \\
                                    & Conv-based                          & 4:1  & 12.50                   \\
                                    & CTC-based                           & 13:1  & 3.85                  \\
                                    & WLQ-former                     & 16:1 & 3.12 \\              
\hline
\end{tabular}
\end{table}

In Table \ref{tab:compression_ratio}, we report compression factors for each
adapter. We also distinguish between the SFMs (SeamlessM4T and Whisper) as their output sequence length is very different and this influences the compression factor, especially for the content-based adapters (CIF-based and CTC-based). 
%configuration of SFM + Adapter. 
For the content-based length adapters,
%configurations, 
we present the estimated compression ratios on the test set, evaluated after training.
Notably, the CIF-based adapter reaches the highest compression ratio (25:1) with Whisper, which corresponds to a $\sim$2 Hz frequency of output vectors. The CTC-based, on the other hand, reduces the output sequence to $\sim$3 Hz using SeamlessM4T while it reduces the sequence to approximately 4 Hz with Whisper. The textual representations of the ground truth transcriptions, as tokenized by the LLM, contain $\sim$2.7 token per second, i.e. slightly longer than those produced by CIF-based and slightly shorter than those produced by the CTC-based and WLQ-former.

\subsection{LLM Decoder}
As LLMs, we select Mistral-7B-Instruct-v0.3,\footnote{\url{https://huggingface.co/mistralai/Mistral-7B-Instruct-v0.3}} and Llama-3.1-8B-Instruct\footnote{\url{https://huggingface.co/meta-llama/Meta-Llama-3.1-8B-Instruct}}. The former is an English-centric model while the latter is multilingual, covering English, German, French, Italian, Portuguese, Hindi, Spanish, and Thai. Moreover, Llama is the most popular choice in previous works \cite{speech_llama,zhang2023tuning,llast}.
Also in this case, our goal is to maximize the difference between the investigated LLMs while keeping the dimension close.

\begin{table}
\centering
\caption{Number of 
%trainable 
parameters for each adapter.
%configuration, separated into those used for length adaptation and those used for modality adaptation.
}
\label{tab:adapter_parameters}
\begin{tabular}{l|ccc}
\hline
\multirow{2}{*}{\textbf{Adapter}} & \multicolumn{3}{c}{\textbf{\# trainable parameters (M)}} \\ \cline{2-4} 
                                  & \multicolumn{1}{c}{\textbf{len. adapt.}} & \multicolumn{1}{c}{\textbf{mod. adapt.}} & \textbf{Total} \\ \hline
%Transformer
Base & \multicolumn{1}{r}{0}                      & 28.35          &             28.35                   \\ \hline
% C-Former (CIF)
Conv-based & \multicolumn{1}{r}{25.20}                  & 28.35         &          53.55                                \\ \hline

CIF-based
& \multicolumn{1}{r}{28.15}                  & 28.35              &          56.50                  \\ \hline

% C-Former (CTC)
CTC-based & \multicolumn{1}{r}{25.20}                  & 28.35        &        53.55                          \\ \hline
% C-Former (Conv)
% Q-Former
WLQ-former & \multicolumn{3}{c}{33.09}                                                                  \\
\hline
\end{tabular}
\end{table}

\begin{table*}[!ht]
\small
\centering
\setlength{\tabcolsep}{2.5pt}
\renewcommand{\arraystretch}{1.05} % Default value: 1
% \caption{Comparison of model performances across different language pairs. The best result for a given language pair under a fixed ⟨audio encoder, text decoder⟩ configuration is \underline{underlined}, with the overall best being also \textbf{bolded}.}
\caption{ASR and ST results on CoVoST test sets. The best result for each (SFM, LLM) configuration is  \underline{underlined}, while the overall best is \textbf{bolded}. The difference with Base is statistically significant ($p<0.05$) unless for scores marked with $^{*}$.}
\label{tab:model_performance}
\begin{tabular}{c|c|l|cccccc|cccccc}
\hline
 \multirow{2}{*}{\textbf{SFM}}   &    \multirow{2}{*}{\textbf{LLM}}                                    &             \multirow{2}{*}{\textbf{Adapter}}     & 
%\multicolumn{6}{c}{\textbf{\textbf{↑ COMET score (CoVoST - ST)}}}                                                          & \multicolumn{6}{c}{\textbf{↓~WER (CoVoST - ASR)}}
\multicolumn{6}{c|}{\textbf{\textbf{ST - COMET ($\uparrow$)}}}                                                          & \multicolumn{6}{c}{\textbf{ASR - WER ($\downarrow$)}}
 \\
\cline{4-15}
                         &   &  & \textbf{en-de}         & \textbf{de-en} & \textbf{es-en} & \textbf{fr-en} & \textbf{it-en} & \textbf{avg}            & \textbf{en} & \textbf{es} & \textbf{fr} & \textbf{it} & \textbf{de} & \textbf{avg}  \\ 
\cline{1-15}
\multirow{10}{*}{\rotatebox{90}{SeamlessM4T}} & \multirow{5}{*}{\rotatebox{90}{Mistral}}   & 
% Transformer
Base 
& \uline{84.94}                  & \uline{84.75}          & \uline{86.65}          & \uline{\textbf{84.71}}          & \uline{\textbf{85.42}}          & \uline{85.29}                    & 6.48        & \uline{6.56}        & \uline{\textbf{9.69}}        & \uline{\textbf{7.8}}         & \uline{\textbf{8.36}}        & \uline{\textbf{7.78}}          \\
                                     &                                      & 
%C-Former (CIF) 
CIF-based
& 84.31                  & 84.33          & 86.31          & 84.32          & 85.07          & 84.87                    & 7.10        & 6.92        & 10.23       & 8.60        & 9.38        & 8.45          \\
                                     &                                      & 
% C-Former (Conv)
Conv-based
& 84.33                  & 84.15          & 86.20          & 84.11          & 84.98          & 84.75                    & 7.53        & 7.83        & 11.38       & 10.07       & 11.44       & 9.65          \\
                                     &                                      & 
% C-Former (CTC) 
CTC-based
& 82.95                  & 82.48          & 85.20          & 82.85          & 83.57          & 83.41                    & 7.94        & 7.90         & 12.51       & 10.31       & 12.29       & 10.19         \\
                                     &                                      & 
% Q-Former [q1] 
WLQ-former 
& 84.67                  & \,\,\,84.71$^{*}$          & \,\,\,86.60$^{*}$          & 84.59          & 85.29          & 85.17                    & \,\,\,\uline{\textbf{6.38}}$^{*}$        & 6.80         & \,\,\,9.83$^{*}$        & 8.05        & \,\,\,8.48$^{*}$        & 7.91          \\ 
\cline{2-15}
                                     & \multirow{5}{*}{\rotatebox{90}{Llama 3.1}} & 
% Transformer 
Base
& 85.12                  & 84.15          & 86.17          & 84.08          & 84.78          & 84.86                    & 7.15        & 7.46        & 10.67       & 9.20         & 9.96        & 8.89          \\
                                     &                                      & 
% C-Former (CIF)
CIF-based
& 84.65 & 83.87          & 85.98          & 83.86          & 84.65 & 84.60   & 7.66        & \,\,\,7.47$^{*}$        & 12.36        & 10.18        & 10.50        & 9.63         \\
                                     &                                      & 
% C-Former (Conv)
Conv-based
& 85.42                  & 84.42          & 86.43          & 84.31          & 85.17          & 85.15                    & \,\,\,7.16$^{*}$        & 7.08        & 10.79$^{*}$       & 8.99        & \,\,\,9.83$^{*}$       & 8.77          \\
                                     &                                      & 
% C-Former (CTC)
CTC-based
& 83.78                  & 82.49          & 85.21          & 82.83          & 83.60          & 83.58                    & 7.95        & 8.04        & 12.17       & 9.94        & 11.22         & 9.90           \\
                                     &                                      & 
% Q-Former [q1]
WLQ-former 
& \uline{\textbf{85.65}}                 & \uline{\textbf{84.84}}  & \uline{\textbf{86.66}}  & \uline{84.68}  & \uline{85.39}  & \uline{\textbf{85.44}}   & \uline{6.62}        & \uline{6.69}        & \uline{9.96}        & \uline{7.97}        & \uline{8.71}        & \uline{7.99}         \\ 
\cline{1-15}
\multirow{12}{*}{\rotatebox{90}{Whisper}}  & \multirow{5}{*}{\rotatebox{90}{Mistral}}   & 
% Transformer
Base
& 78.98                  & 81.38          & 84.79          & 81.63          & 82.69          & 81.89                    & \uline{11.37}       & \uline{7.57}        & \uline{12.81}       & \uline{10.14}       & \uline{10.88}       & \uline{10.55}        \\
                                     &                                      & 
% C-Former (CIF)
CIF-based
& 77.79                  & 80.35          & 84.11          & 80.79          & 81.83          & 80.99                    & 12.57       & 8.45        & 14.24       & 12.32       & 13.09       & 12.13         \\
                                     &                                      & 
% C-Former (Conv)
Conv-based
& 78.73                  & 81.26          & \,\,\,84.72$^{*}$           & \,\,\,81.52$^{*}$           & 82.58          & 81.76                    & 11.78       & 7.60         & 13.23       & 10.67       & 11.52       & 10.96         \\
                                     &                                      & 
% C-Former (CTC)
CTC-based
& 75.56                  & 76.53          & 81.75          & 78.33          & 78.55          & 78.14                    & 14.69       & 10.63       & 17.15       & 15.09       & 16.50        & 14.81        \\
%                                     &                                      & 
%% Q-Former [q4]
%WLQ-former [q4]
%& 78.73                  & 81.39          & 84.65          & 81.45          & 82.70          & 81.78                    & 11.79       & \uline{7.53}        & 13.53       & 10.36       & 11.66       & 10.97        &   \\
                                     &                                      & 
% Q-Former [q1]
WLQ-former
& \,\,\,\uline{79.07}$^{*}$                 & \,\,\,\uline{81.44}$^{*}$         & \uline{84.92}          & \,\,\,\uline{81.68}$^{*}$         & \uline{82.92}          & \uline{82.00}                   & 11.82       & 8.21        & 13.60        & 15.77       & 12.55       & 12.39         \\ 
\cline{2-15}
                                     & \multirow{5}{*}{\rotatebox{90}{Llama 3.1}} & 
% Transformer
Base
& 80.43                  & 82.15          & 85.21          & 82.33          & 83.06          & 82.64                    & \uline{9.90}         & \uline{\textbf{6.33}}        & \uline{11.27}       & \uline{8.52}        & 9.09        & \uline{9.02}         \\
                                     &                                      & 
% C-Former (CIF)
CIF-based
& 78.32                  & 78.94          & 82.51          & 80.09          & 80.27          & 80.02                    & 12.82       & 8.53       & 14.31       & 12.80       & 13.53      & 12.40        \\
                                     &                                      & 
% C-Former (Conv)
Conv-based
& \uline{80.84}                  & \uline{82.57}          & \uline{85.49}          & \uline{82.60}          & \uline{83.51}          & \uline{83.00}                    & \,\,\,\uline{9.90}$^{*}$        & 6.46        & \,\,\,11.49$^{*}$       & \,\,\,8.75$^{*}$        & \,\,\,\uline{9.00}$^{*}$         & 9.12          \\
                                     &                                      & 
% C-Former (CTC) 
CTC-based
& 76.47                  & 73.80          & 80.16          & 77.19          & 76.59          & 76.84                    & 14.02       & 10.98       & 17.55       & 16.29       & 17.21       & 15.21         \\
%                                     &                                      & 
% Q-Former [q4]
%WLQ-former [q4]
%& \uline{81.48}          & \uline{83.14}  & \uline{85.76}  & \uline{82.90}  & \uline{83.91}  & \uline{83.44}            & \uline{9.88}        & 6.37        & 11.71       & \uline{8.51}        & \uline{9.03}        & 9.10         &   \\
                                     &                                      & 
% Q-Former [q1]
WLQ-former
& 79.95                  & 81.56          & 84.88          & 81.56          & 82.89          & 82.17                    & 11.98       & 7.90         & 14.52       & 11.10        & 12.84       & 11.67        \\
\hline
\end{tabular}
\end{table*} 

\section{Experimental Settings}
\label{sec:exp_set}

%\noindent \textbf{Training settings.} 
We trained all our models on CoVoST 2 \cite{wang2020covost} using English, German, Spanish, French, and Italian as source languages and German and English as target languages, and on MuST-C \cite{mustc} using German, French, Italian, Spanish as target languages. The datasets for ASR were obtained by taking the audios in the source languages and their corresponding transcriptions for CoVoST, while for MuST-C we took audios and associated transcriptions from the English-German partition.
Our training process consisted of 28,000 steps over 2 epochs. This was done using either 4 NVIDIA GH200 96GB GPUs with a micro-batch size of 10 samples and 4 gradient accumulation steps, or 8 NVIDIA A100 40GB GPUs with a micro-batch size of 10 samples and 2 gradient accumulation steps. This approach allowed us to maintain a total batch size of 160 samples for both configurations. We took the last checkpoint as the final model for evaluation in all cases since early experiments showed better results compared to averaging the last ones. A single training run took approximately 24 hours on both described systems.
We used AdamW as the optimizer and a cosine scheduler for the learning rate with $10^{-4}$ as the peak value and a linear warmup of 840 steps. The hyperparameters of all the Transformer layers in the adapters are the same as in the BERT\textsubscript{BASE} model.\footnote{\url{https://huggingface.co/google-bert/bert-base-uncased}} 
%Models have been evaluated keeping
To ensure a fair comparison, we kept the number of modality adaptation parameters constant across the different architectures, with the exception of the WLQ-Former where a distinction between length and modality adaptation parameters cannot be made. The exact number of trainable parameters for each configuration, with a separation of length adaptation and modality adaptation parameters where possible, is reported in Table \ref{tab:adapter_parameters}.

%\noindent \textbf{Evaluation.} 
Results are reported as COMET\footnote{Model: \url{https://huggingface.co/Unbabel/wmt22-comet-da}.}~\cite{rei-etal-2020-comet} and WER\footnote{Computed using \texttt{jiwer} (\url{https://pypi.org/project/jiwer/}).} for, respectively, ST and ASR on the CoVoST 2 test sets. Statistical significance is computed using bootstrap resampling \cite{koehn-2004-statistical}.

\section{Results}
\label{sec:results}

In Table \ref{tab:model_performance}, we report the ASR and ST results of the SFM+LLM architectures with the various combinations of SFMs, LLMs, and length adapters introduced in Section \ref{sec:method}. We did not benchmark inference times as autoregressive text decoding dominates over non-autoregressive encoding of (adapted) audio features.

As previous works mostly rely on BLEU \cite{papineni-etal-2002-bleu}, we highlight that our Whisper, Base adapter, and Llama model -- a basic configuration adopted by previous SFM+LLM works -- scores 28.7 BLEU on CoVoST en-de (the most popular language direction), a significantly higher result than the 25.1 BLEU of Qwen-Audio \cite{chu2023qwen}, one of the best performing SFM+LLM solutions. This confirms the soundness of our experimental settings and the competitiveness of our results.
%Our experiments are competitive with those reported in the literature, as the widely adopted SFM+LLM configuration of previous work based on Whisper, Base adapter (no compression), and Llama scores 28.7 BLEU on CoVoST en-de (the most popular language direction) against 25.1 of Qwen-Audio \cite{chu2023qwen}.

First, we observe that the choice of \textbf{the SFM is the most critical factor} in terms of downstream performance. This is shown not only by the fact that, in each configuration, the version equipped with SeamlessM4T outperforms the counterpart with Whisper on both tasks (ASR and ST) on average, but also by the average improvement of more than 2 COMET on ST and more than 1 WER on ASR of the best SeamlessM4T configurations (with Llama and the WLQ-former or Base adapters) over the best Whisper ones (with Llama and the Conv-based adapter for ST and Llama and Base for ASR).
%results significantly vary when a different SFM is adopted. The best combinations equipped with SeamlessM4T as the speech encoder achieve an average improvement of about 3.3 and 2 COMET on ST and 2.8 and 1.4 WER on ASR, respectively using Mistral and Llama as LLM, compared to the Whisper speech encoder.
Instead, the choice of the LLM is less critical, as demonstrated by the small gap ($<$0.2 on both ASR and ST) between the best configuration with Llama (Seamless as SFM and WLQ-former as adapter) and that with Mistral (SeamlessM4T as SFM and Base as adapter). Moreover, the best results on average in ST are obtained with Llama, while the best ones in ASR with Mistral.\\
\indent Second, results clearly show that \textbf{there is no one-size-fits-all solution for the adapter}. Interestingly, the LLM plays an important role in the choice of the adapter. With Mistral, the Base adapter generally yields the best results, even though the WLQ-former is competitive, especially in ST. 
With Llama, instead, the best adapter varies with the SFM used. While WLQ-former is the best option with SeamlessM4T, the Conv-based and Base adapters emerge with Whisper, with the former being the best in ST and close in ASR, where differences among the two are almost always not statistically significant.
Across all SFM and LLM configurations and tasks, the Base adapter always ranks first or second except for Llama+SeamlessM4T in ST, where it is third. Moreover, content-based length adapters consistently underperform other strategies. Together with the observation that there is no clear trend of the results with respect to the compression factor, these insights suggest that reducing the length mismatch between textual and speech representations 
%(which are typically $\sim$10 times longer) 
is not critical for the quality of the outputs. However, reducing the speech sequence length lowers computational costs, making length adapters still a useful module to consider.\\
\indent All in all, our results demonstrate the need for experimenting in different settings in terms of SFM and LLM when comparing adapter solutions, as improvements in one specific scenario may not generalize. In addition, LLMs shown to be robust to input sequences of very different lengths, as the Base adapter, which does not compress the speech sequence, and the WLQ-former, which has high compression factors (16 with Whisper), achieve competitive scores in most settings.

\section{Conclusions}
\label{sec:conclusion}

This work systematically analyzed the importance and design of the various building blocks that compose speech-to-text models by connecting an SFM encoder and an LLM through an adapter. To this aim, we compared all the combinations of 2 SFMs, 2 LLMs, and 5 adapters, which mostly differ in their length reduction module. With comprehensive experiments covering two tasks--ASR and ST--and 5 language directions, our results demonstrate that the choice of the SFM is the most critical factor influencing downstream performance. We also established that there is no one-size-fits-all solution for the length adapter, as the optimal choice varies depending on the specific combination of SFM and LLM.
%, with different adapters performing best under different conditions. 
Notably, the Base and WLQ-former adapters, which feature very different compression factors, demonstrate strong performance across tasks, suggesting that reducing sequence length mismatch between speech and text is less crucial than previously assumed.
%These findings highlight the importance of evaluating novel adapter solutions across multiple SFM and LLM configurations, as performance improvements in one setting may not generalize.

\section{Acknowledgments}

This paper has received funding from the European Union’s Horizon research and innovation programme under grant agreement No 101135798, project Meetween (My Personal AI Mediator for Virtual MEETtings BetWEEN People).

We gratefully acknowledge Poland's high-performance Infrastructure PLGrid ACC Cyfronet AGH for providing computer facilities and support within computational grant no PLG/2024/016971.

Marco Gaido was supported by the PNRR project FAIR - Future AI Research (PE00000013), under the NRRP MUR program funded by the NextGenerationEU.

This work has been carried out while Pierfrancesco Melucci was enrolled in the Italian National Doctorate on Artificial Intelligence run by Sapienza University of Rome in collaboration with Università di Roma Tre.

\newpage

\bibliographystyle{IEEEtran}
\bibliography{conference_101719_interspeech}

\end{document}